\documentclass[10pt,conference]{IEEEtran}
\IEEEoverridecommandlockouts
\usepackage{cite}
\usepackage{amsmath,amssymb,amsfonts}
\usepackage{algorithmic}
\usepackage{graphicx}
\usepackage{textcomp}
\usepackage{xcolor}
\usepackage{multirow}
\usepackage{makecell}
\usepackage{tikz}
\usepackage{pgfplots}
\usepgfplotslibrary{groupplots}
\pgfplotsset{compat=1.18}
\usepackage[capitalise,noabbrev]{cleveref} 

\usepackage{tikz}
\usetikzlibrary{patterns}
\usepackage{pgfplots}
\pgfplotsset{compat=1.18}
\usetikzlibrary{external}

\newcolumntype{L}[1]{>{\raggedright\let\newline\\\arraybackslash\hspace{0pt}}m{#1}}
\newcolumntype{C}[1]{>{\centering\let\newline\\\arraybackslash\hspace{0pt}}m{#1}}
\newcolumntype{R}[1]{>{\raggedleft\let\newline\\\arraybackslash\hspace{0pt}}m{#1}}

\def\BibTeX{{\rm B\kern-.05em{\sc i\kern-.025em b}\kern-.08em
    T\kern-.1667em\lower.7ex\hbox{E}\kern-.125emX}}

\usepackage[caption=false]{subfig}
\usepackage[nolist]{acronym}
\usepackage{balance}

\makeatletter
\newcommand{\linebreakand}{%
  \end{@IEEEauthorhalign}
  \hfill\mbox{}\par
  \mbox{}\hfill\begin{@IEEEauthorhalign}
}
\makeatother

\begin{acronym}[TDMA]
	\acro{AUC-ROC}{Area under the Receiver Operating Characteristic Curve}
    \acro{CNN}{Convolutional Neural Network}
    \acro{DAMS}{Difficulty Adaptive Margin Scheduler}
    \acro{EER}{Equal Error Rate}
    \acro{FPR}{False Positive Rate}
    \acro{GMM}{Gaussian Mixture Model}
    \acro{MLP}{Multilayer Perceptron}
    \acro{ROC}{Receiver Operating Characteristic}
    \acro{DML}{Distance Metric Learning}
    \acro{LFW}{Labelled Faces in the Wild}
    \acro{LITM}{Learning Incremental Triplet Margin}
\end{acronym}

\begin{document}

\title{Maintaining Difficulty: A Margin Scheduler for Triplet Loss in Siamese Networks Training 
\thanks{This work has been supported by the Brazilian National Council for Scientific and Technological Development (CNPq) -- Grants 444820/2024-8 and 405511/2022-1.}}

\author{\IEEEauthorblockN{Roberto Sprengel Minozzo Tomchak}
\IEEEauthorblockA{\textit{Departamento de Informática} \\
\textit{Universidade Federal do Paraná}\\
Curitiba, Brazil \\
robertotomchak@ufpr.br}
\and
\IEEEauthorblockN{Oge Marques}
\IEEEauthorblockA{\textit{College of Engineering and Computer Science} \\
\textit{Florida Atlantic University}\\
Boca Raton, FL, U.S.A \\
omarques@fau.edu}
\and
\IEEEauthorblockN{Lucas Garcia Pedroso}
\IEEEauthorblockA{\textit{Departamento de Matemática} \\
\textit{Universidade Federal do Paraná}\\
Curitiba, Brazil \\
lucaspedroso@ufpr.br}
\and

\linebreakand

\IEEEauthorblockN{Luiz Eduardo Oliveira}
\IEEEauthorblockA{\textit{Departamento de Informática} \\
\textit{Universidade Federal do Paraná}\\
Curitiba, Brazil \\
luiz.oliveira@ufpr.br}
\and
\IEEEauthorblockN{Paulo Lisboa de Almeida}
\IEEEauthorblockA{\textit{Departamento de Informática} \\
\textit{Universidade Federal do Paraná}\\
Curitiba, Brazil \\
paulorla@ufpr.br}
}


\maketitle

\begin{abstract}
The Triplet Margin Ranking Loss is one of the most widely used loss functions in Siamese Networks for solving \acf{DML} problems. This loss function depends on a margin parameter $\mu$, which defines the minimum distance that should separate positive and negative pairs during training. In this work, we show that, during training, the effective margin of many triplets often exceeds the predefined value of $\mu$, provided that a sufficient number of triplets violating this margin is observed. This behavior indicates that fixing the margin throughout training may limit the learning process. Based on this observation, we propose a margin scheduler that adjusts the value of $\mu$ according to the proportion of easy triplets observed at each epoch, with the goal of maintaining training difficulty over time. We show that the proposed strategy leads to improved performance when compared to both a constant margin and a monotonically increasing margin scheme. Experimental results on four different datasets show consistent gains in verification performance. Our trained models and source code are available at github.com/robertotomchak/maintaining-difficulty.
\end{abstract}


\section{Introduction}\label{sec:intro}

\begin{figure}[htbp]
    \centering
    \includegraphics[width=1.0\linewidth]{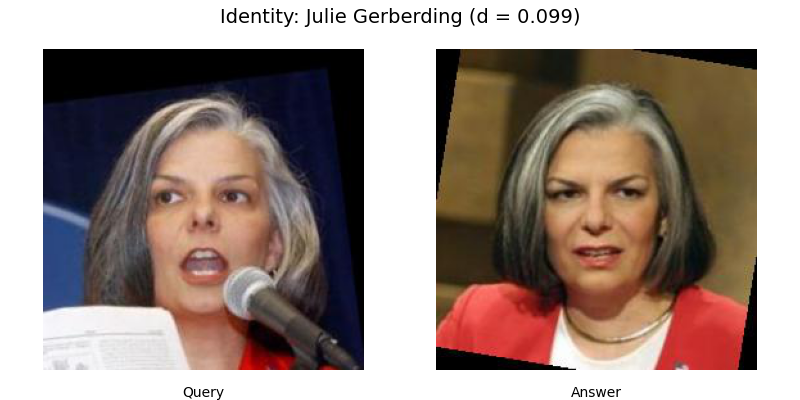}
    \caption{To identify the person in the query image (left), the image is compared with all samples in a database, and the identity corresponding to the smallest distance is selected (right) as the answer. Images from \ac{LFW} \cite{huang2008labeled}.}
    \label{fig:siamese_classification}
\end{figure}

\acf{DML} tasks focus on training a model to transform samples into feature embeddings such that similar samples are mapped close to each other, while dissimilar samples are mapped further apart. This is a common approach for problems in which the goal is to assess whether two samples belong to the same class \cite{schroff2015facenet}.
Typical applications include verifying whether two images belong to the same person \cite{schroff2015facenet}, finding the closest match of an image within a database, as illustrated in Figure~\ref{fig:siamese_classification} \cite{schroff2015facenet}, tracking objects by comparing them across multiple images \cite{ribas2024using}, or transforming a classification problem into a comparison-based one by matching an instance against labeled samples \cite{li2022survey}.

A common approach to train such models is to employ a loss function that takes triplets of samples into account \cite{BMVC2016_119}. These triplets are defined as $T = (A, P, N)$, where $A$ is the anchor sample, $P$ is a positive sample belonging to the same class as $A$, and $N$ is a negative sample belonging to a different class. The objective is to enforce the distance between the embeddings of $A$ and $P$ to be smaller than the distance between the embeddings of $A$ and $N$ by a given margin, thus clustering similar samples while keeping dissimilar ones apart.

Formally, let $f(X)$ denote the embedding of sample $X$. The positive distance is defined as $\delta_+(T) = d(f(A), f(P))$ and the negative distance as $\delta_-(T) = d(f(A), f(N))$, where $d$ is a distance function, typically the Euclidean distance \cite{BMVC2016_119}. The desired condition is $\delta_+(T) + \mu < \delta_-(T)$, where $\mu \geq 0$ denotes the desired margin. Triplets that do not satisfy this inequality must trigger an update in the model parameters. This behavior is commonly enforced using the Triplet Margin Ranking Loss~\cite{BMVC2016_119}, which is defined as $l(T) = \max(0, \mu + \delta_+(T) - \delta_-(T))$. Triplets that satisfy the margin constraint yield a zero loss and are referred to as easy, while those that violate the constraint, producing a positive loss, are referred to as hard~\cite{BMVC2016_119, schroff2015facenet}.

We define the effective margin of a triplet $T$ as $\hat{\mu}(T) = \delta_-(T) - \delta_+(T)$. From the definitions above, a triplet is easy if and only if $\hat{\mu}(T) > \mu$, and hard otherwise. Consequently, triplets whose effective margin is greater than or equal to the desired margin $\mu$ yield a zero loss.

One may incorrectly assume that, since easy triplets produce a zero loss, their effective margins remain unchanged over time. Under this assumption, triplets would tend to cluster around the desired margin $\mu$, and there would be no incentive for the model to further increase the separation of easy triplets. In this work, however, we show that this intuition is incorrect. Despite yielding zero loss, the effective margins of easy triplets do increase during training. We argue that this behavior arises because updates triggered by hard triplets reshape the embedding space, indirectly affecting the distances of easy~triplets.

Moreover, most existing works adopt a constant value of $\mu$ throughout training \cite{schroff2015facenet, li2022survey, BMVC2016_119}. We show that gradually increasing the margin over time is beneficial. Based on this observation, we propose two margin schedulers: a linear scheduler, which increases the margin $\mu$ linearly over training epochs, and an adaptive scheduler, called \ac{DAMS}, which increases $\mu$ when the proportion of easy triplets reaches a predefined threshold. The adaptive scheduler can be interpreted as an attempt to maintain the difficulty level of the training triplets over time.

Experiments conducted on four well-known datasets from different domains, namely \acs{LFW}, CelebA, CUB-200-2011, and Stanford Cars, show that networks trained using the proposed margin schedulers achieve better performance compared to a constant margin $\mu$. These results indicate that the proposed approaches are not domain-specific. In summary, the main contributions of this work are as follows:

\begin{itemize}
    \item We analyze how the effective margins of easy triplets evolve during training, showing that they are implicitly updated together with hard triplets, leading to improved embeddings for both types of triplets.
    \item We propose margin scheduling strategies that progressively increase the desired margin during training, enabling a larger number of effective updates and improving the overall training process.
\end{itemize}

The remainder of this paper is organized as follows. \cref{sec:relatedWorks} reviews related work. \cref{sec:proposed_method} analyzes how the effective margin achieved during training often exceeds the predefined value of $\mu$ and uses this observation to motivate the proposed adaptive method. The experimental protocol is described in \cref{sec:exp_protocol}, and the experimental results are presented in \cref{sec:experiments}. Finally, conclusions and directions for future work are discussed in \cref{sec:conclusion}.

\section{Related Works}\label{sec:relatedWorks}

Training networks for comparison-related tasks using loss functions that enforce a minimum margin $\mu$ between negative pairs is a common approach, adopted in works such as~\cite{wang2014learning, schroff2015facenet, BMVC2016_119, boutros2022self, zhang2025low}. Some studies have explored alternatives to using a constant margin in the triplet loss or in related loss functions, such as the contrastive loss. These works generally follow one of two main directions: assigning a different margin to each triplet or batch \cite{ha2021deep,kim2022adaface} or modifying the margin value as training progresses \cite{zhang2019learning, kim2022adaface, moonrinta2025AdaptiveTriplet}.

An example of the first direction is the work of \cite{ha2021deep}, which proposes an adaptive margin in the context of rating datasets. In this approach, the similarity between two samples is combined with their relative ratings to define the margin. Samples with similar ratings are associated with smaller margins, while larger rating differences lead to larger margins. Similar ideas are explored in \cite{semedo2019cross, moonrinta2025improving}. Although promising, these methods rely on additional information beyond class labels, such as explicit ratings between training pairs.

The authors of \cite{zhang2019learning} propose a method in which the margin is increased during training. Their approach, called \ac{LITM}, starts from a small base margin and divides training into multiple stages. At each stage, the margin is increased by a predefined amount, which can vary across stages. Each margin increment is treated as a hyperparameter. In their experiments, three stages are used with margins set to $(4, 7, 10)$, although the paper does not provide clear guidelines on how these values should be selected in practice.

In \cite{kim2022adaface}, the authors introduce AdaFace, an adaptive loss function that adjusts the margin according to image quality. Since image quality is difficult to estimate directly, the method uses the norm of the image embedding within the batch as a proxy, under the assumption that higher norms correspond to higher-quality images.

More recently, \cite{moonrinta2025AdaptiveTriplet} proposed an adaptive triplet loss that employs different margins for positive and negative pairs. The authors also highlight the difficulty of choosing fixed margin values and propose a temporal strategy in which margins are updated using an exponential moving average. In this method, the updated margin is computed as a weighted combination of the previous moving average and the mean pairwise distances observed in the current batch.

While existing approaches often focus on specific scenarios, such as exploiting additional dataset information or adapting the margin based on batch-level statistics, our work aims to provide a more general strategy. We propose updating the margin only when the learning problem becomes sufficiently easy, based on an explicit analysis of training difficulty across~epochs.

\section{Proposed Method}\label{sec:proposed_method}

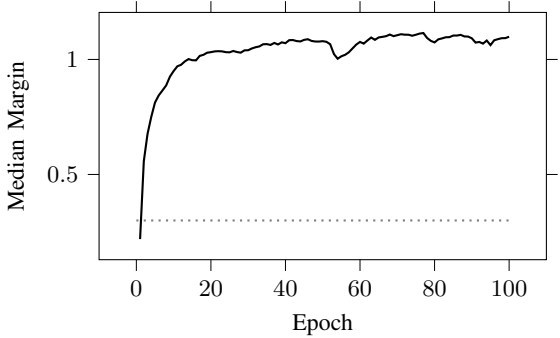
\begin{figure}[htbp]
    \centering
    \begin{tikzpicture}
\begin{axis}[
    width=0.85\linewidth,
    height=0.55\linewidth,
    xlabel={Epoch},
    ylabel={Median Margin},
    tick style={black},
    tick align=outside,
    axis line style={black},
    label style={font=\small},
    tick label style={font=\small},
]
\addplot[
    thick,
    black
]
table[
    x=epoch,
    y=median_margin,
    col sep=comma
]{tikz/median_margin.csv};

\addplot[
    gray,
    dotted,
    thick
] coordinates {
    (0,0.3)
    (100,0.3)
};
\end{axis}
\end{tikzpicture}
    \caption{Median effective margin of triplets at each training epoch on the LFW dataset.}
    \label{fig:median_margin}
\end{figure}

As discussed in Section~\ref{sec:intro}, updates to the embedding space affect both easy and hard triplets. Consider Figure~\ref{fig:median_margin}, which shows the median value of $\hat{\mu}$ (effective margin) computed over all training triplets across 100 training epochs.
This experiment follows the protocol described in Section~\ref{sec:exp_protocol}, using the LFW dataset and a fixed margin $\mu = 0.3$.
If the embeddings of easy triplets remained unchanged during training, one would expect the median effective margin to saturate close to $0.3$. Instead, as training progresses, the median effective margin continues to increase well beyond this value, reaching approximately $1.1$ in the final epoch.

\begin{figure}[htpb]
  \centering
  \subfloat[Epoch 1]{\includegraphics[trim={0.8cm 0 1.2cm 1.2cm},clip,width=0.24\textwidth]{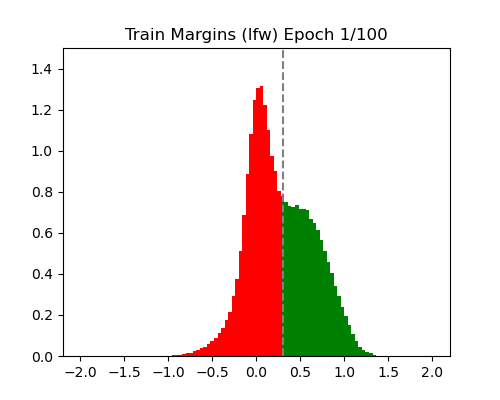}}
  \hfill
  \subfloat[Epoch 5]{\includegraphics[trim={0.8cm 0 1.2cm 1.2cm},clip,width=0.24\textwidth]{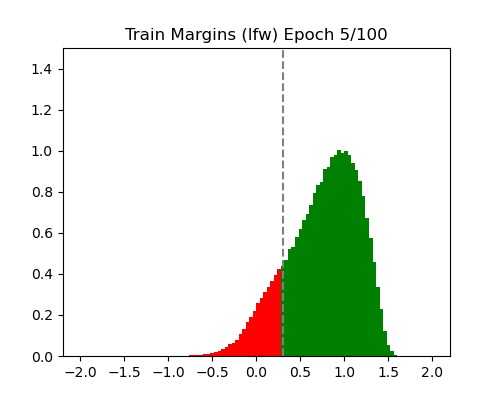}}\\
  \subfloat[Epoch 50]{\includegraphics[trim={0.8cm 0 1.2cm 1.2cm},clip,width=0.24\textwidth]{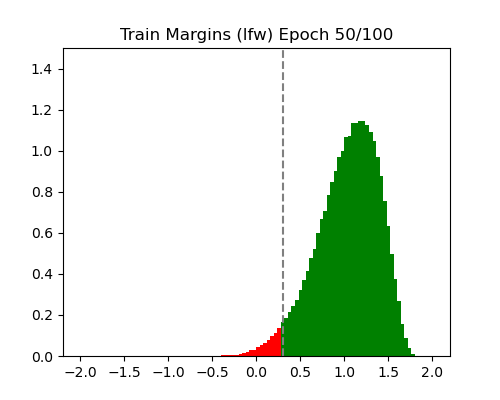}}
  \hfill
  \subfloat[Epoch 100]{\includegraphics[trim={0.8cm 0 1.2cm 1.2cm},clip,width=0.24\textwidth]{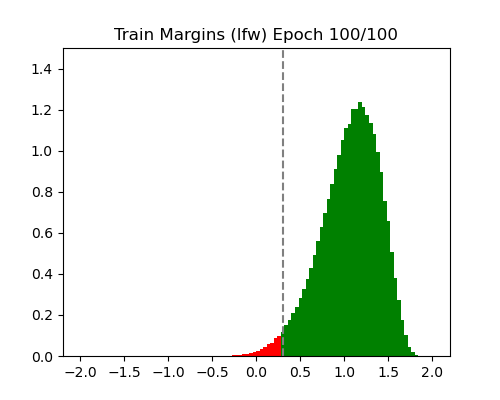}}
  \caption{Histogram of effective margins across training epochs using the LFW dataset. Values below $\mu = 0.3$ correspond to hard triplets (red), while values above correspond to easy triplets (green).}
  \label{fig:hists_margin}
\end{figure}

This provides evidence that the effective margins of triplets keep increasing during training, even after they become easy. We observed similar behavior across all other datasets evaluated in this work. The histograms shown in \cref{fig:hists_margin} further support this observation, indicating that the distribution of effective margins progressively shifts to values larger than $\mu$ as training advances.

Our hypothesis is that, while the model enforces the desired margin $\mu$ for hard triplets, it learns increasingly discriminative features that also improve the separation of easy triplets. For instance, consider a face recognition model. Early in training, the model may rely on simple attributes, such as hair color, which are sufficient to separate some triplets. As training progresses and the model is forced to handle harder triplets, such as faces with similar hair color, it must learn more detailed features, such as eye color or facial structure. As these features are learned, triplets that already exhibited large effective margins can become even better separated.

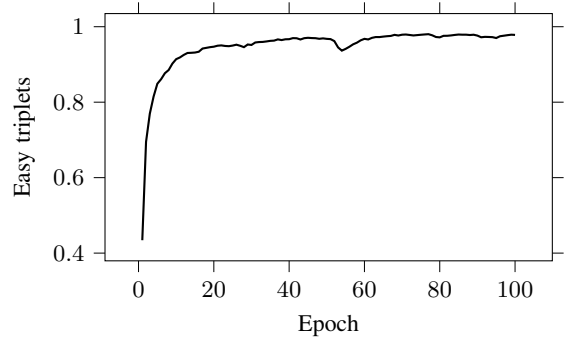
\begin{figure}[htbp]
    \centering
    \begin{tikzpicture}
\begin{axis}[
    width=0.85\linewidth,
    height=0.55\linewidth,
    xlabel={Epoch},
    ylabel={Easy triplets},
    tick style={black},
    tick align=outside,
    axis line style={black},
    label style={font=\small},
    tick label style={font=\small},
]
\addplot[
    thick,
    black
]
table[
    x=epoch,
    y=easy_percent,
    col sep=comma,
    header=true
]{tikz/easy_percent.csv};
\end{axis}
\end{tikzpicture}
    \caption{Proportion of easy triplets per epoch on the LFW dataset.}
    \label{fig:easy_percent}
\end{figure}

However, using a small margin $\mu$ may lead to a rapidly decreasing number of hard triplets, causing the model to focus on a limited subset of the data. This subset may include a disproportionate number of outliers or noisy samples, potentially hindering further improvements \cite{moonrinta2025AdaptiveTriplet}. As an example, Figure~\ref{fig:easy_percent} shows the proportion of easy triplets per epoch under the same experimental protocol. This proportion grows rapidly, causing the number of hard triplets to decrease quickly. When 80\% of triplets are easy, which occurs at epoch 4, the training loss decreases by an average of 2.3\% per epoch. In contrast, when 95\% of triplets are easy, which occurs at epoch 22, the average decrease drops to 0.61\% per epoch.

These observations indicate that while a small margin is beneficial during the early stages of training, its effectiveness diminishes as the number of easy triplets becomes too large. To address this issue, we propose dynamically adjusting the margin $\mu$ in order to maintain a sufficient number of hard triplets throughout training. Specifically, we evaluate two margin scheduling strategies: a Linear scheduler and an Adaptive~scheduler. The Linear scheduler starts from an initial margin $\mu_0$ and increases it linearly at each epoch by a constant increment $s_l$.

The Adaptive scheduler also starts from an initial margin $\mu_0$. Whenever the proportion of easy triplets in a given epoch exceeds a predefined threshold $t$, the margin is increased by a fixed increment $s_a$. This threshold $t$ can be interpreted as a target level of task difficulty. We refer to this adaptive strategy as \acf{DAMS}.

\section{Experimental Protocol}\label{sec:exp_protocol}

In this section, we present the experimental protocol adopted in this work. Since our goal is to evaluate the behavior of the proposed margin schedulers in comparison to a constant margin strategy, we do not focus on large models or extensive data preprocessing to maximize absolute performance. Instead, we employ a relatively small model (\cref{sec:training}) and use the datasets as provided.

\subsection{Datasets}\label{sec:datasets}

We evaluate our approach on the following datasets: \ac{LFW}, CelebA, CUB-200-2011, and Stanford Cars. \ac{LFW} consists of face images from multiple individuals and presents a high imbalance in the number of samples per class. Approximately 70\% of identities have only one image, while some individuals have more than 200 images. CelebFaces Attributes (CelebA) is a similar dataset composed exclusively of celebrity faces. In contrast to \ac{LFW}, CelebA is more balanced, with more than 92\% of identities having at least five images. CUB-200-2011 is a fine-grained dataset of bird images, where each class corresponds to a distinct bird species. Finally, Stanford Cars contains images of cars, where classes are typically defined by the tuple (manufacturer,~model,~year).

We split the datasets following protocols commonly adopted in the literature, ensuring that the classes in the training and test sets are disjoint. CUB-200-2011 and Stanford Cars are split evenly, with 50\% of the classes used for training and 50\% for testing \cite{ji2023siamese,do2019theoretically,harwood2017smart}. \ac{LFW} is split according to its recommended protocol, resulting in approximately 70\% of the classes for training and 30\% for testing \cite{huang2008labeled}. For CelebA, we adopt the same splitting strategy used for \ac{LFW}. \cref{table:datasets} summarizes the datasets used in our experiments. Throughout this work, we refer to a class as a single identity, such as a person or a bird species, which may contain multiple images. Classes containing only one image are removed, as they cannot be used to form valid triplets or positive pairs.

\begin{table}[htpb]
\centering
\begin{tabular}{lrrrr}\hline
\multirow{2}{*}{Dataset} & \multicolumn{2}{c}{Train} & \multicolumn{2}{c}{Test}\\
& \# Classes & \# Images & \# Classes & \# Images \\\hline
\ac{LFW} & 1,184 & 6,671 & 496 & 2,493 \\
CelebA & 7,094 & 141,859 & 3,039 & 60,696 \\
CUB-200-2011 & 100 & 5,899 & 100 & 5,897 \\
Stanford Cars & 98 & 8,109 & 98 & 8,076 \\
\hline
\end{tabular}
\caption{Train and test splits of the datasets used in the experiments.}
\label{table:datasets}
\end{table}

\subsection{Metrics}\label{sec:metrics}

We evaluate performance using two metrics: the \ac{AUC-ROC} and Recall@$k$, with $k \in \{1, 2, 4, 8\}$. The \ac{AUC-ROC} metric is computed using pairs of samples, with the goal of determining whether two samples belong to the same class \cite{schroff2015facenet}. Recall@$k$ is a common metric in \ac{DML} problems \cite{do2019theoretically}, which measures whether a sample can be correctly identified among its $k$ nearest neighbors.

To compute the \ac{AUC-ROC}, we generate pairs in the test set as follows. For each class $c_k$, with $k = 1,\ldots,K$, we randomly sample one positive pair $(A_k, P_k)$, where both samples belong to class $c_k$ and $A_k \neq P_k$, and one negative pair $(A_k, N_k)$, where $N_k$ does not belong to class $c_k$. This procedure ensures a balanced set of positive and negative pairs while covering all classes. The total number of pairs is therefore twice the number of classes, making it proportional to the dataset size.

To compute Recall@$k$, each sample $a$ in the test set is temporarily removed from the dataset. We then retrieve the $k$ nearest samples to $a$ among the remaining test samples and verify whether at least one of them belongs to the same class as $a$.

\subsection{Model and Training}\label{sec:training}

We use a \ac{CNN} pretrained on the ImageNet \cite{imagenet} dataset, which is a common practice in Siamese network training \cite{li2022survey,do2019theoretically}. Specifically, we adopt EfficientNetB0 \cite{tan2019efficientnet} as the backbone network. Following \cite{schroff2015facenet}, the backbone is followed by three fully connected layers: two hidden layers with 512 and 256 neurons, respectively, and an output layer with 128 neurons, followed by a $L_2\text{-normalization layer}$ \cite{schroff2015facenet}. As a result, the generated embedding vectors are 128-dimensional and live in a hypersphere of radius one and centered at the origin.

All models are trained using the Adam optimizer with a learning rate of 0.001. Training is performed for 100 epochs with a batch size of 64 images. In-triplet hard negative mining, defined in \cite{BMVC2016_119}, is employed in all experiments.

\subsection{Tested Margin Schedulers}

We evaluate the two proposed margin scheduling strategies. For the Linear scheduler, we initialize the margin with $\mu_0 = 0.0$ and increase it linearly by adding $s_l = 0.01$ at the end of each epoch. This results in a final margin of $1.0$ at epoch 100. For the proposed \ac{DAMS} scheduler, we also use $\mu_0 = 0.0$ and a step size of $s_a = 0.01$. If the margin were increased at every epoch, this would make the adaptive scheduler equivalent to the linear one, allowing for a fair comparison between the two~strategies.

We set the threshold for the adaptive scheduler to $t = 0.95$, based on the observations in Section~\ref{sec:proposed_method}, where the training loss was shown to decrease significantly less after this proportion of easy triplets was reached. Finally, we compare both schedulers against a constant margin baseline using a fixed margin value of $\mu = 0.3$.

\section{Experiments and Results}\label{sec:experiments}

The experiments presented in this section are reported as the average of three executions. We first present the results considering all datasets in \cref{sec:mainExperiments}. Then, in \cref{sec:ablation}, we perform a brief hyperparameter analysis to evaluate the influence of the proposed parameters, considering the \ac{LFW}~dataset.

\subsection{Comparison of Margin Schedulers}\label{sec:mainExperiments}

Table \ref{table:metrics} presents the results obtained on each dataset. We start by analyzing the linear scheduler, which, despite not achieving the best overall performance, outperformed the constant-margin approach in most scenarios. These results indicate that updating the margin over time is beneficial, which is consistent with previous works, such as \cite{moonrinta2025AdaptiveTriplet,zhang2019learning}.

Our proposed \ac{DAMS} adaptive method achieved the best results in most scenarios, with the linear scheduler outperforming it only in terms of \ac{AUC-ROC} on the Stanford Cars dataset. This result suggests that increasing the desired margin $\mu$ only when the problem becomes sufficiently simple can be more effective than updating it at a constant rate.

\begin{table*}[htpb] \centering \caption{Metrics Results with each Method in the Datasets} \begin{tabular}{llrrrrrr} \hline Dataset & Margin & AUC & Recall@1 & Recall@2 & Recall@4 & Recall@8 \\ \hline \multirow{3}{*}{\ac{LFW}} & Constant & 0.956 $\pm$0.01 & 0.273 $\pm$0.04 & 0.385 $\pm$0.04 & 0.500 $\pm$0.04 & 0.620 $\pm$0.04 \\ & \ac{DAMS} & \textbf{0.966 $\pm$0.00} & \textbf{0.395 $\pm$0.02} & \textbf{0.500 $\pm$0.02} & \textbf{0.605 $\pm$0.01} & \textbf{0.703 $\pm$0.01} \\ & Linear & 0.963 $\pm$0.00 & 0.364 $\pm$0.02 & 0.476 $\pm$0.02 & 0.578 $\pm$0.02 & 0.679 $\pm$0.02 \\ \hline \multirow{3}{*}{CelebA} & Constant & 0.953 $\pm$ 0.00 & 0.137 $\pm$ 0.00 & 0.211 $\pm$ 0.01 & 0.303 $\pm$ 0.01 & 0.411 $\pm$ 0.01 \\ & \ac{DAMS} & \textbf{0.956 $\pm$ 0.00} & \textbf{0.217 $\pm$ 0.02} & \textbf{0.309 $\pm$ 0.02} & \textbf{0.411 $\pm$ 0.02} & \textbf{0.519 $\pm$ 0.02} \\ & Linear & 0.940 $\pm$ 0.00 & 0.101 $\pm$ 0.00 & 0.161 $\pm$ 0.00 & 0.243 $\pm$ 0.00 & 0.347 $\pm$ 0.00 \\ \hline \multirow{3}{*}{CUB-200-2011} & Constant & 0.897 $\pm$0.02 & 0.270 $\pm$0.03 & 0.406 $\pm$0.02 & 0.550 $\pm$0.02 & 0.683 $\pm$0.03 \\ & \ac{DAMS} & \textbf{0.899 $\pm$0.02} & \textbf{0.379 $\pm$0.01} & \textbf{0.515 $\pm$0.00} & \textbf{0.642 $\pm$0.00} & \textbf{0.755 $\pm$0.00} \\ & Linear & 0.894 $\pm$0.00 & 0.334 $\pm$0.01 & 0.475 $\pm$0.02 & 0.613 $\pm$0.01 & 0.734 $\pm$0.01 \\ \hline \multirow{3}{*}{Stanford Cars} & Constant & 0.933 $\pm$0.00 & 0.329 $\pm$0.04 & 0.474 $\pm$0.05 & 0.621 $\pm$0.05 & 0.742 $\pm$0.04 \\ & \ac{DAMS} & 0.941 $\pm$0.01 & \textbf{0.439 $\pm$0.02} & \textbf{0.576 $\pm$0.02} & \textbf{0.699 $\pm$0.01} & \textbf{0.796 $\pm$0.01} \\ & Linear & \textbf{0.953 $\pm$0.01} & 0.432 $\pm$0.01 & 0.572 $\pm$0.01 & 0.698 $\pm$0.01 & 0.800 $\pm$0.00 \\ \hline \multirow{3}{*}{Average} & Constant & 0.928 & 0.856 & 0.242 & 0.355 & 0.476\\ & \ac{DAMS} & \textbf{0.937} & \textbf{0.865} & \textbf{0.346} & \textbf{0.464} & \textbf{0.577}\\ & Linear & 0.934 & 0.861 & 0.292 & 0.402 & 0.516 \\\hline \end{tabular} \label{table:metrics} \end{table*}

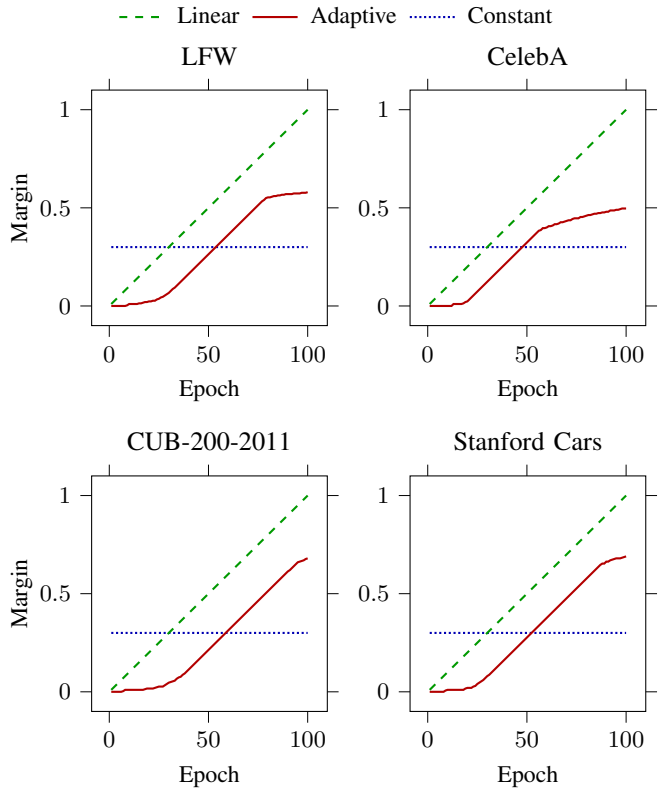
\begin{figure}[htbp]
    \centering
%
%
\begin{tikzpicture}

\begin{axis}[%
hide axis,
xmin=-0.48,
xmax=7.5,
ymin=3.0,
ymax=9.0,
ylabel={$f_2$},
axis x line*=bottom,
axis y line*=left,
legend style={legend cell align=left,align=left,draw=black,legend columns=3, column sep=2.5pt, draw=none, font=\small}
]

\addlegendimage{thick, green!60!black, dashed}
\addlegendentry{Linear};

\addlegendimage{thick, red!70!black}
\addlegendentry{Adaptive};

\addlegendimage{thick, blue!70!black, densely dotted}
\addlegendentry{Constant};

\end{axis}
\end{tikzpicture}%
\\
    \hspace{-0.1cm}\begin{tikzpicture}
\begin{groupplot}[
    group style={
        group size=2 by 2,
        horizontal sep=1.1cm,
        vertical sep = 2cm
    },
    width=0.26\textwidth,
    height=0.26\textwidth,
    xlabel={Epoch},
    ylabel={Margin},
    tick style={black},
    tick align=outside,
    axis line style={black},
    label style={font=\small},
    tick label style={font=\small},
    legend style={
        draw=none,
        font=\small,
        at={(0.5,-0.45)},
        anchor=north,
        legend columns=3
    }
]

\nextgroupplot[
    title={LFW},
    ylabel style={yshift=-4pt}
]
\addplot[thick, blue!70!black, densely dotted]
table[x=epoch, y=lfw_hard, col sep=comma, header=true]{tikz/margins_methods.csv};

\addplot[thick, red!70!black]
table[x=epoch, y=lfw_adaptive, col sep=comma, header=true]{tikz/margins_methods.csv};

\addplot[thick, green!60!black, dashed]
table[x=epoch, y=lfw_linear, col sep=comma, header=true]{tikz/margins_methods.csv};

\nextgroupplot[
    title={CelebA},
    ylabel=\empty
]
\addplot[thick, blue!70!black, densely dotted]
table[x=epoch, y=celebA_hard, col sep=comma, header=true]{tikz/margins_methods.csv};

\addplot[thick, red!70!black]
table[x=epoch, y=celebA_adaptive, col sep=comma, header=true]{tikz/margins_methods.csv};

\addplot[thick, green!60!black, dashed]
table[x=epoch, y=celebA_linear, col sep=comma, header=true]{tikz/margins_methods.csv};

\nextgroupplot[
    title={CUB-200-2011},
    ylabel style={yshift=-4pt}
]
\addplot[thick, blue!70!black, densely dotted]
table[x=epoch, y=birds_hard, col sep=comma, header=true]{tikz/margins_methods.csv};

\addplot[thick, red!70!black]
table[x=epoch, y=birds_adaptive, col sep=comma, header=true]{tikz/margins_methods.csv};

\addplot[thick, green!60!black, dashed]
table[x=epoch, y=birds_linear, col sep=comma, header=true]{tikz/margins_methods.csv};

\nextgroupplot[
    title={Stanford Cars},
    ylabel=\empty
]
\addplot[thick, blue!70!black, densely dotted]
table[x=epoch, y=cars_hard, col sep=comma, header=true]{tikz/margins_methods.csv};

\addplot[thick, red!70!black]
table[x=epoch, y=cars_adaptive, col sep=comma, header=true]{tikz/margins_methods.csv};

\addplot[thick, green!60!black, dashed]
table[x=epoch, y=cars_linear, col sep=comma, header=true]{tikz/margins_methods.csv};

\end{groupplot}
\end{tikzpicture}
    \caption{Margin evolution for each method across datasets.}
    \label{fig:margins_methods}
\end{figure}

In \cref{fig:margins_methods}, we show the margin evolution throughout training for all datasets. In the CelebA dataset, the \ac{DAMS} scheduler exhibits a slow margin increase, with the curve becoming nearly horizontal around epoch 60 at approximately $\mu = 0.5$. This suggests that the linear scheduler likely increased the margin too aggressively, which may explain its poorer performance on this dataset, being the only case where the constant-margin method outperformed it.

We hypothesize that the linear method probably increased the margin too aggressively. This is corroborated by the large number of classes in the CelebA dataset, with approximately 3,000 identities in the test set, which may require a lower margin to fit all embeddings in the 128-dimensional space. In contrast, the Stanford Cars and CUB-200-2011 datasets contain fewer classes, around 100 in the test set, and can therefore accommodate higher margins, which may explain the higher final margins achieved by the \ac{DAMS} scheduler.

The results on \ac{LFW} align with this analysis, as its number of classes is closer to that of CelebA, resulting in a similar final margin. Importantly, the proposed \ac{DAMS} scheduler is able to adapt to these differences automatically, achieving appropriate margins across datasets while using the same hyperparameter configuration.

\subsection{Hyperparameter Analysis}\label{sec:ablation}

As discussed in \cref{sec:proposed_method}, the \ac{DAMS} scheduler has three hyperparameters: the initial margin $\mu_0$, the threshold of easy triplets $t$, and the step size $s_a$. The optimal values for these parameters are not obvious. To analyze their influence, we evaluate different combinations using the \ac{LFW} dataset, following the experimental protocol described in \cref{sec:exp_protocol}. To do this, we test the following values:
\\
\\
\begin{itemize}
    \item $\mu_0 \in \{0.0, 0.3\}$.
    \item $t \in \{50\%, 80\%, 85\%, 90\%, 95\%, 99\%\}$.
    \item $s_a \in \{0.01, 0.02, 0.05, 0.10\}$.
\end{itemize}

This results in $2 \cdot 6 \cdot 4 = 48$ configurations. We focus on the Recall@1 and \ac{AUC-ROC} metrics. Table \ref{table:ablation_results} reports the ten best and ten worst configurations, sorted in decreasing order of Recall@1.

\begin{table}[htpb] \centering \caption{Results for Different Parameters in the Adaptive Method} \begin{tabular}{lrrrr} \hline $\mu_0$ & $t$ & $s_a$ & Recall@1 & AUC\\ \hline \multicolumn{4}{c}{Best Configurations}\\\hline 

0.0 & 99\% & 0.02 & 0.411 $\pm$ 0.02 & 0.967 $\pm$ 0.00\\
0.0 & 95\% & 0.01 & 0.403 $\pm$ 0.01 & 0.963 $\pm$ 0.01\\
0.0 & 99\% & 0.05 & 0.387 $\pm$ 0.03 & 0.965 $\pm$ 0.01\\
0.0 & 95\% & 0.02 & 0.383 $\pm$ 0.00 & 0.964 $\pm$ 0.00\\
0.0 & 99\% & 0.01 & 0.380 $\pm$ 0.02 & 0.966 $\pm$ 0.00\\
0.0 & 90\% & 0.01 & 0.373 $\pm$ 0.04 & 0.966 $\pm$ 0.00\\
0.0 & 95\% & 0.05 & 0.366 $\pm$ 0.02 & 0.964 $\pm$ 0.00\\
0.0 & 99\% & 0.1 & 0.359 $\pm$ 0.03 & 0.968 $\pm$ 0.01\\
0.0 & 80\% & 0.01 & 0.351 $\pm$ 0.01 & 0.963 $\pm$ 0.00\\
0.0 & 85\% & 0.01 & 0.345 $\pm$ 0.01 & 0.958 $\pm$ 0.00\\

\hline \multicolumn{4}{c}{Worst Configurations}\\\hline 

0.3 & 50\% & 0.01 & 0.180 $\pm$ 0.01 & 0.914 $\pm$ 0.01\\
0.3 & 80\% & 0.05 & 0.176 $\pm$ 0.01 & 0.919 $\pm$ 0.00\\
0.0 & 80\% & 0.1 & 0.173 $\pm$ 0.00 & 0.916 $\pm$ 0.01\\
0.3 & 80\% & 0.1 & 0.159 $\pm$ 0.02 & 0.914 $\pm$ 0.01\\
0.0 & 50\% & 0.02 & 0.136 $\pm$ 0.01 & 0.841 $\pm$ 0.02\\
0.3 & 50\% & 0.02 & 0.108 $\pm$ 0.00 & 0.846 $\pm$ 0.02\\
0.0 & 50\% & 0.05 & 0.106 $\pm$ 0.01 & 0.817 $\pm$ 0.01\\
0.3 & 50\% & 0.05 & 0.099 $\pm$ 0.03 & 0.807 $\pm$ 0.01\\
0.0 & 50\% & 0.1 & 0.085 $\pm$ 0.02 & 0.801 $\pm$ 0.02\\
0.3 & 50\% & 0.1 & 0.072 $\pm$ 0.02 & 0.801 $\pm$ 0.01\\

\hline \end{tabular} \label{table:ablation_results} \end{table}

As can be observed in  \cref{table:ablation_results}, the best configurations are predominantly associated with high values of $t$, while the worst results are obtained with lower values. Only three of the top-performing configurations use $t < 0.95$, and none use $t < 0.8$. This suggests that training benefits from maintaining a small proportion of hard triplets, indicating that the dataset contains limited noise or that noisy samples do not significantly affect the loss. This behavior is consistent with the analysis presented in \cref{sec:proposed_method}.

No clear trend is observed regarding the step size $s_a$, indicating that the proposed method is relatively robust to this parameter within the tested range. Finally, these results should be interpreted with caution, as the analysis is limited to the \ac{LFW} dataset. A more comprehensive investigation across additional datasets is left for future work.

\section{Conclusions}\label{sec:conclusion}

In this work, we show that even when easy triplets satisfy the desired margin $\mu$ in the Triplet Margin Ranking Loss and therefore produce zero loss, the embedding updates driven by hard triplets substantially influence these easy triplets. As a result, their effective margins increase over the course of training. This observation highlights a discrepancy between the fixed target margin enforced by the loss function and the evolving geometry of the embedding space.

Motivated by this behavior, we propose the \acf{DAMS}, which adapts the desired margin $\mu$ during training. Instead of increasing the margin according to a predefined schedule, \ac{DAMS} relies on the proportion of easy triplets observed at each epoch, increasing $\mu$ only when the learning problem becomes sufficiently easy.

Experimental results on four datasets covering distinct fine-grained and face recognition tasks show that \ac{DAMS} consistently outperforms both a constant-margin strategy and a monotonically increasing margin scheduler. The same set of hyperparameters was used across all datasets, suggesting that the proposed method generalizes well without dataset-specific tuning. While we do not claim that \ac{DAMS} is universally optimal for all metric learning scenarios, these results indicate that adapting the margin based on training difficulty is an effective strategy.

We hypothesize that the improved performance of \ac{DAMS} over linear scheduling arises from its ability to prevent over-constraining the embedding space. In contrast to methods that increase the margin independently of the data distribution, \ac{DAMS} implicitly accounts for factors such as the number of classes and embedding dimensionality, reducing the risk of enforcing margins that are incompatible with the embedding-space.

Nevertheless, \ac{DAMS} has some limitations. First, because margin updates depend on reaching a target proportion of easy triplets, the maximum achievable margin growth is inherently bounded for a given step size. In scenarios where faster margin expansion is beneficial, such as the Stanford Cars dataset, this behavior may slow convergence. Second, the method introduces three hyperparameters that require selection. However, we showed a group of hyperparameters that seem to work well in a range of scenarios ($t=0.95, \mu_0 = 0$ and $s_a = 0.01$).

In future works, we will better investigate these hyperparameters Also, the graphs in Figure \ref{fig:margins_methods} show that the margin tends to saturate in the \ac{DAMS} method at some point, which indicates we might be able to use it as a stopping criterion.

\bibliographystyle{IEEEtran}
\bibliography{IEEEabrv,bibliography}

\end{document}